\documentclass[12pt]{article}
\usepackage[utf8]{inputenc}
\usepackage[T1]{fontenc}
\usepackage{geometry}
\usepackage{authblk}
\usepackage{amssymb}
\usepackage{multirow}
\usepackage{amsmath}
\usepackage{graphicx}
\usepackage{caption}
\usepackage{subcaption}
\usepackage[table,xcdraw]{xcolor}
\usepackage{bm} 
\usepackage{adjustbox}
\usepackage{algorithm}
\usepackage{algorithmic}
\usepackage{array}
\usepackage{cancel}
\usepackage{xfrac}
\usepackage{tikz}
\usepackage{multirow}
\usepackage{float}
\newcolumntype{C}[1]{>{\centering\arraybackslash}m{#1}}
\usepackage{graphicx}
\usepackage{subcaption}
\usepackage{caption}
\usepackage{tabularx}
\usepackage[numbers]{natbib}
\usepackage{makecell}
\usepackage{url}

\providecommand{\keywords}[1]{%
  \noindent\textbf{\textit{Keywords---}} #1
}

\title{Canine Clinical Gait Analysis for Orthopedic and Neurological Disorders:\\ An Inertial Deep-Learning Approach}

\author[1]{Netta Palez}
\author[2]{Léonie Straß}
\author[2]{Sebastian Meller}
\author[2]{Holger Volk}
\author[3]{Anna Zamansky}
\author[1]{Itzik Klein}

\affil[1]{The Hatter Department of Marine Technologies, University of Haifa, Haifa, Israel}
\affil[2]{Department of Small Animals Medicine and Surgery, University of Veterinary Medicine Hannover, Germany}
\affil[3]{Department of Information Systems, University of Haifa, Haifa, Israel}

\date{} 

\begin{document}

\maketitle

\begin{abstract}
Canine gait analysis using wearable inertial sensors is gaining attention in veterinary clinical settings, as it provides valuable insights into a range of mobility impairments. Neurological and orthopedic conditions cannot always be easily distinguished even by experienced clinicians. The current study explored and developed a deep learning approach using inertial sensor readings to assess whether neurological and orthopedic gait could facilitate gait analysis. Our investigation focused on optimizing both performance and generalizability in distinguishing between these gait abnormalities. Variations in sensor configurations, assessment protocols, and enhancements to deep learning model architectures were further suggested. Using a dataset of 29 dogs, our proposed approach achieved 96\% accuracy in the multiclass classification task (healthy/orthopedic/neurological) and 82\% accuracy in the binary classification task (healthy/non-healthy) when generalizing to unseen dogs. Our results demonstrate the potential of inertial-based deep learning models to serve as a practical and objective diagnostic and clinical aid to differentiate gait assessment in orthopedic and neurological conditions.
\end{abstract}

\vspace{6mm}


\newpage

\section{Introduction}

Canine gait analysis is crucial for understanding numerous health conditions. For example, neurological gait disorders can manifest in diverse ways, including ataxia and paresis \cite{paluvs2014neurological}. Ataxia refers to impaired coordination and may result from dysfunction in the peripheral nerves or spinal cord (proprioceptive ataxia), the vestibular system (vestibular ataxia), or the cerebellum (cerebellar ataxia) \cite{paluvs2014neurological}. Paresis, characterized by reduced voluntary movement and muscle strength, contrasts with plegia, which denotes a complete loss of voluntary movement. Lameness is more commonly associated with pain due to orthopedic conditions \cite{garosi2013neurological}, but mild cases of paraparesis \cite{park2024characterization} and diseases affecting the nerve root(s) can cause the same clinical presentation \cite{park2024characterization}. Orthopedic lameness is often marked by weight shifting, shortened steps, and a reduced stance phase \cite{witte2011investigation}, accompanied by an elongated, compensating stride in the contralateral limb \cite{paluvs2014neurological}. \\
Traditionally, canine gait analysis is performed by visual inspection and subjective assessment by a trained clinician. While overt lameness is often recognizable through such observation, more subtle gait abnormalities may go undetected without objective tools, making accurate diagnosis particularly challenging in early or mild cases \cite{kerwin2021assessment}. Besides the inability of the human eye to capture the complexity of each component of movement \cite{weyer2007reliability}, evaluation still depends on personal experience and education \cite{waxman2008relationship}. Moreover, studies have shown that even with objective scoring systems, inter, and intra-observer reliability remains low \cite{quinn2007evaluation, lee2015preliminary}. This highlights the need for more objective methods of gait assessment. In recent years, such approaches have gained growing interest in veterinary gait analysis. Notable examples include video-based kinematic analysis \cite{fu2010evaluation} and kinetic analysis using force plates or pressure mats \cite{gillette2008recent}. Although these methods provide greater objectivity and precision, their use is limited by the high cost of equipment and the need for specialized expertise. \\
The use of wearable inertial measurement units (IMUs) provides an attractive alternative. IMUs are usually equipped with accelerometers and gyroscopes, which can be used to study complex dynamics not captured by kinematic or kinetic analysis \cite{hayati2019analysis, piche2022validity}. Moreover, their compact size, low cost, and fast 3D motion acquisition, make wearable IMU sensors suitable for free-roaming canine gait studies. A growing body of research has explored the use of IMUs in canine gait and locomotion analysis, including the characterization of gait patterns \cite{jenkins2018automatic, ladha2017gaitkeeper, altermatt2023extraction, clark2014evaluation}, detection of lameness \cite{rhodin2017inertial}, analysis of locomotion \cite{gerencser2013identification}, and monitoring of body orientation \cite{britt2011embedded}. Several studies  have focused on canine gait abnormalities caused by specific diseases, including dystrophin-deficiency \cite{barthelemy2011longitudinal}, ligament ruptures \cite{pillard20123d}, lack of locomotor-cardiac coupling \cite{simmons1997lack} and ataxia \cite{engelsman2022measurement}. \\
In most of these studies, a single IMU sensor was mounted on a dog’s main body, such as the neck, back, or sternum, which only allows assessment of general body movement. Recently, Zhang et al. \cite{zhang2022four} investigated a four-limb IMU platform, validating its performance against a commercial pressure-sensor-based walkway system. While this multi-sensor setup offers more precise gait measurements, it may compromise the dog’s comfort and potentially alter natural walking behaviour.
Recent advancements in hardware and computational efficiency have demonstrated the effectiveness of deep learning (DL) methods in real-time applications, including image processing, signal processing, and natural language processing, by leveraging their ability to handle nonlinear problems \citep{lecun2015deep, goodfellow2016deep, shinde2018review, mahrishi2020machine}. Consequently, DL methods have started to be incorporated into inertial navigation algorithms, opening the door for using DL models with inertial sensors in general \cite{cohen2024inertial, golroudbari2023generalizable}. While deep learning has shown promising results in human gait analysis \cite{shah2025gait, santos2019low}, its application to canine gait assessment remains relatively underexplored. The integration of DL with inertial sensor data has the potential to enhance early detection of orthopedic and neurological conditions in dogs, yet existing studies have not fully leveraged this capability. \\
In this study, we propose a practical inertial deep-learning approach for canine clinical gait analysis for orthopedic and neurological disorders.  We employ wearable inertial sensors placed at multiple locations on the dog’s body to investigate the differentiation between orthopedic and neurological clinical conditions, which often present with similar gait abnormalities.
To this end, we explore the following research questions:

\begin{enumerate}
    \item What is the optimal sensor configuration - both in terms of number and placement - for distinguishing between orthopedic and neurological conditions?
    \item What is the most effective assessment protocol - specifically, which activities should the dog perform, such as walking or running - for optimally distinguishing between orthopedic and neurological conditions?
    \item How can deep-learning techniques be applied to inertial data using the sensor configuration and assessment protocols above to accurately differentiate between orthopedic and neurological conditions?
    \item How well do the developed deep learning models generalize to unseen dogs?
    \item How can the developed deep learning approach performance be further enhanced?
\end{enumerate}

To answer the above research questions, we employ a unique dataset consisting of 29 dogs including 17 healthy dogs, 6 dogs with orthopedic disorders and 6 dogs with neurological disorders. We develop a simple yet effective end-to-end neural network capable of multiclass classification of different clinical conditions. \\
The rest of the paper is organized as follows: Section 2 introduces our proposed model, covering both its architecture and the training process. Section 3 provides an overview of the recorded dataset, a key component of this research. Section 4 presents our experimental results. Finally, Section 5 gives the conclusions of this research.

\section{Proposed Approach} \label{proposed_approach}

Our approach leverages a simple, yet efficient, neural network to learn patterns from accelerometer and gyroscope readings and determine if the dog is healthy or has orthopedic or neurological abnormalities. In this section, we detail the architecture of our proposed network, including its layer composition, the unique aspects of our model, and the parameters used during training. We then describe the research methodology.

\subsection{Network Architecture} \label{Net_Arch}

\begin{figure}[h!]
\centering
\captionsetup{justification=centering}
\includegraphics[scale=0.4]{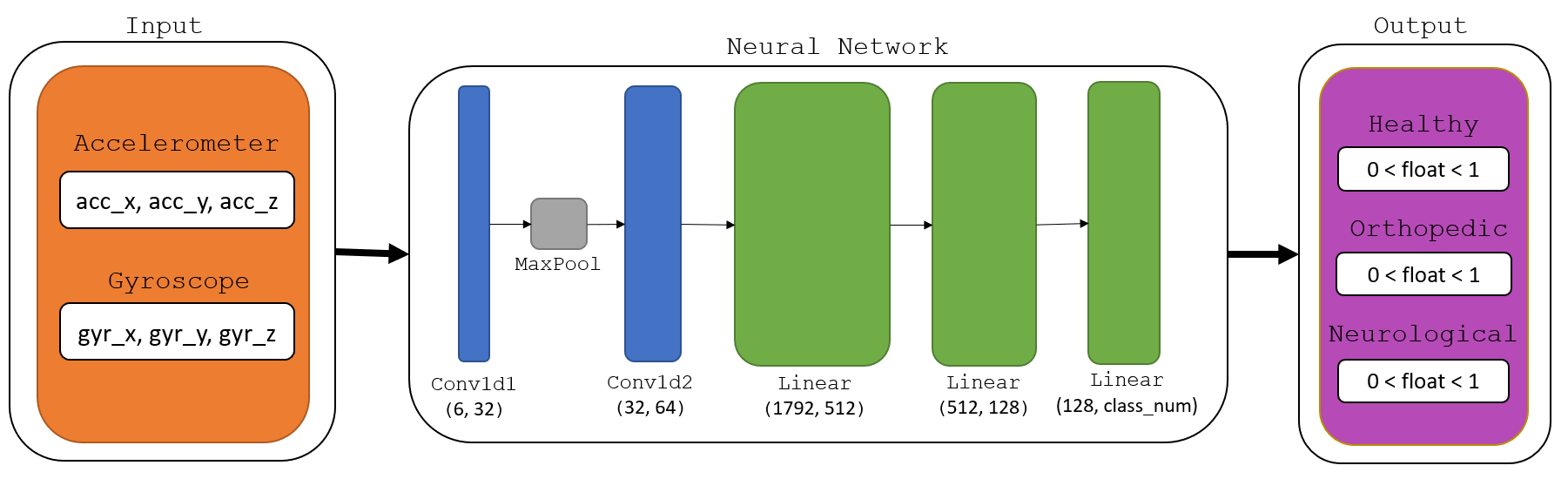}
\caption{Our proposed neural network for the classification task.}
\label{fig:archi}
\end{figure}

Our proposed architecture is illustrated in Figure \ref{fig:archi}. It consists of two convolutional neural network (CNN) layers followed by three fully connected (FC) layers. The CNN layers incorporate max pooling, while the rectified linear unit (ReLU) activation function is applied to all layers except the final one, which uses log-softmax. To mitigate overfitting observed during development, dropout is employed as a regularization technique. Next we provide a detailed mathematical description of these components.

A FC layer is defined by \cite{prince2023understanding}:

\begin{equation}
\begin{aligned}
    \boldsymbol{z}_i^{(k)} = \sum_{j=1}^{n_{k-1}} \mathbf{\Omega}_{ij}^{(k)} \boldsymbol{a}_{j}^{(k-1)} + \boldsymbol{b}_i^{(k)}
    \label{eq:fc}
\end{aligned}
\end{equation}

where $\mathbf{\Omega}_{ij}^{(k)}$ is the weight of the $i$ neuron in the $k$ layer, associated with the output of the $j$ neuron in the $(k-1)$ layer ($\boldsymbol{a}_{j}^{(k-1)}$), $\boldsymbol{b}_i^{(k)}$ represents the bias in layer $k$ of the $i$ neuron, and $n_{k-1}$ represents the number of neurons in the $(k-1)$ layer. \newline
The ReLU activation function is \cite{agarap2018deep}:

\begin{equation}
\begin{aligned}
    \text{ReLU}(\boldsymbol{z}_i^{(k)}) = max(0, ~ \boldsymbol{z}_i^{(k)})
    \label{eq:relu}
\end{aligned}
\end{equation}

\vspace{2mm}

By assuming a $m_1 \times m_2$ filter (or kernel) the convolutional layer can be written as follows:

\begin{equation}
\begin{aligned}
    \mathbf{C}_{ij}^{(k)} = \sum_{\alpha=0}^{m_1} \sum_{\beta=0}^{m_2}
    \mathbf{\Omega}_{\alpha\beta}^{(l)} \boldsymbol{a}_{(i+\alpha)(j+\beta)}^{(k-1)}
    + b^{(l)}
    \label{eq:CNN}
\end{aligned}
\end{equation}

where $\mathbf{\Omega}_{\alpha\beta}^{(l)}$ is the weight in the $(\alpha, \beta)$ position of the $l$ convolutional layer, $b^{(l)}$ represents the bias of the $l$ convolutional layer and $\boldsymbol{a}_{(ij)}^{(k-1)}$ is the output of the preceding layer.

Given an input time series sample representing raw inertial measurements:

\begin{equation}
    \boldsymbol{x} = \left[{f_x, f_y, f_z, \omega_x, \omega_y, \omega_z}\right] ^T \in \mathbb{R}^{6 \times n}
\end{equation}

where $f_x, f_y, f_z$ are the accelerometer readings along the x, y, and z axes, respectively, $\omega_x, \omega_y, \omega_z$ are the gyroscope readings along the x, y, and z axes, respectively, and n is the window size.

The first stage of the model aims for feature extraction, applies two one-dimensional CNN layers:

\begin{equation}
    \boldsymbol{h}_1 = \text{ReLU}\left(\mathbf{W}_0 \boldsymbol{x} + \boldsymbol{b}_0 \right)
\end{equation}

where $\mathbf{W}_0 \in \mathbb{R}^{6 \times 32}$ is the initial weight matrix and $\boldsymbol{b}_0$ is the initial bias vector. In the same manner, the second convolutional layer follows a similar transformation:

\begin{equation}
    \boldsymbol{h}_2 = \text{ReLU}\left(\mathbf{W}_1 \boldsymbol{h}_1 + \boldsymbol{b}_1 \right),
\end{equation}

where $\mathbf{W}_1 \in \mathbb{R}^{32 \times 64}$ represents the kernel weights of the second convolutional layer.

After feature extraction, the resulting feature map is flattened into a vector $ \boldsymbol{z} \in \mathbb{R}^{d}$, where $d=1792$ is the total number of extracted features. This vector is passed through three FC layers:

\begin{equation}
    \boldsymbol{h}_3 = \text{ReLU}\left(\mathbf{W}_2 \boldsymbol{z} + \boldsymbol{b}_2 \right),
\end{equation}

\begin{equation}
    \boldsymbol{h}_4 = \text{ReLU}\left(\mathbf{W}_3 \boldsymbol{h}_3 + \boldsymbol{b}_3 \right),
\end{equation}

\begin{equation}
    \boldsymbol{\hat{y}} = \text{LogSoftmax}\left(\mathbf{W}_4 \boldsymbol{h}_4 + \boldsymbol{b}_4 \right).
\end{equation}

Here, $\mathbf{W}_2 \in \mathbb{R}^{1792 \times 512}$, $\mathbf{W}_3 \in \mathbb{R}^{512 \times 128}$, and $\mathbf{W}_4 \in \mathbb{R}^{128 \times N}$ are the weight matrices of the FC layers, with the respective hidden layer sizes, and $N=3$ is the number of output classes. The $\text{LogSoftmax}$ function ensures numerical stability when computing the loss.

\subsection{Training Process}
Our training process was across 60 epochs (was chosen after longer trials that converted until 50 epochs). 
The negative log-likelihood (NLL) loss function was used in the training process. It measures how closely our model predictions align with the ground truth (GT) labels. The NLL is defined by \cite{prince2023understanding}: 

\begin{equation}
\begin{aligned}
    L_\text{NLL} =\frac{1}{n}\sum_{i=1}^{n} \big(y_i \log \hat{y}_{i} + (1-y_i)\log(1-\hat{y}_{i}) \big)
    \label{eq:nl_eq}
\end{aligned}
\end{equation}

where $y_i$ is the GT value of the $i^{th}$ sample, $\hat{y}_{i}$ is the predicted value of the $i^{th}$ sample, and n is the number of samples in the corresponding window. \newline
We employ the adaptive moment estimation (Adam) for the optimization process. It is a powerful optimizer that combines the strengths of two other well-known techniques - momentum and RMSprop - to deliver a powerful method for adjusting the learning rates of parameters during training \cite{bock2019proof}. In our model we use Adam with small weight decay of 0.0001 and the learning rate initialization was set to 0.0001 as well. We performed the training process 60 times and used the best model for the testing phase.

\subsection{Methodology} \label{methodology}

To explore canine clinical gait analysis for orthopedic and neurological disorders, this study utilized a descriptive and exploratory methodology, combining different setups to provide a multi-faceted perspective. To this end, we distinguish between the following characteristics: 
\begin{itemize}
    \item  \textbf{Sensor locations}: Four different sensor locations are examined: head, tail, neck(collar) and, all (all the three sensors). In that manner, we can examine how gait features vary between the inertial readings from different locations.
    \item  \textbf{Assessment dynamics}:  We examine two different types of dog dynamics during the recording:  walking and trotting. The motivation is to examine if some features are more dominant in different dynamics allowing for better classification accuracy. 
    \item  \textbf{Classification task}: We consider three practical classification tasks for the problem at hand, namely:
        \begin{enumerate}
            \item  \textbf{Multi}: Classification to multiple classes healthy/orthopedic/neurological.
            \item \textbf{Binary}: Classification to binary classes healthy/non-healthy (orthopedic \\+ neurological).
            \item \textbf{Diagnosis}: Classification to diagnostic classes orthopedic/neurological.
        \end{enumerate}
    \item \textbf{Generalization}: For our baseline models we use a train/test approach based on a random split of the dataset. However, to investigate the generalizability of the developed models to new unseen dogs, we use a stricter leave-one-out (LOO) train/test split, as discussed in [27]. This implies that in each iteration, the model is trained on all dogs except one, which is then used for testing. This process is repeated for each dog, and the aggregated performance metrics are recorded.
\end{itemize}

The chosen methodology is aimed to provide both breadth and depth in understanding the complex relationships examined.

\section{The Dataset} \label{subsec:dataset}

\subsection{Data Collection and Preprocessing}
\begin{figure}[h!]
\centering
\captionsetup{justification=centering}
\includegraphics[scale=0.8]{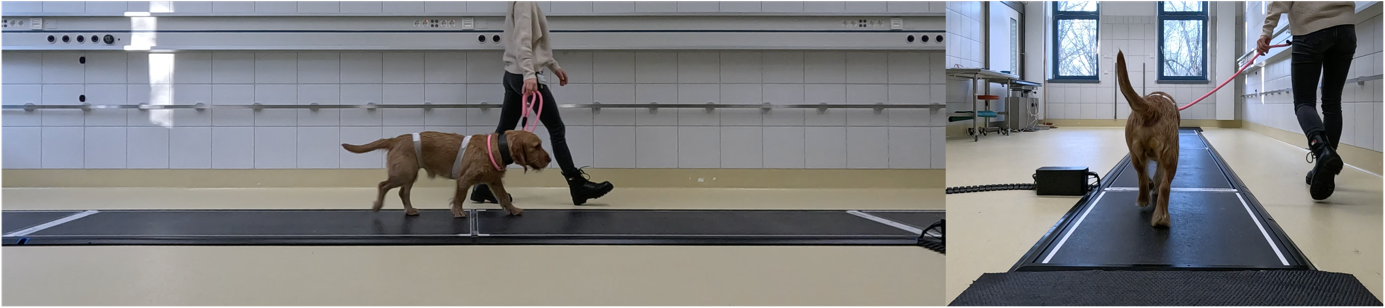}
\caption{Experimental setup in the gait laboratory.}
\label{fig:walking}
\end{figure}

Experimental data were collected at the Department for Small Animal Medicine and Surgery of the University of Veterinary Medicine Hanover, Germany. The study was conducted in a controlled laboratory environment, comprising an 8.75-meter-long testing area. Within this space, dogs simultaneously traversed a pressure-sensing walkway, which was extended at both ends with a sensor-free running surface of the same texture and a non-slip rubber mat, ensuring a smooth transition onto and off the measurement area (Figure \ref{fig:walking}). Dogs were led on a leash while ensuring that the handler consistently remained on the same side of the walkway, resulting in each dog being guided from both the left and right side over the course of the trials. Prior to data acquisition, dogs were permitted to ambulate freely within the testing space to facilitate environmental habituation and minimize potential stress-induced variability. \newline

\begin{table}[h!]
\centering
\resizebox{\textwidth}{!}{%
\begin{tabular}{|c|c|c|c|c|c|}
\hline
Group & Male & \begin{tabular}[c]{@{}c@{}}Male-\\ neutered\end{tabular} & Female & \begin{tabular}[c]{@{}c@{}}Female-\\ neutered\end{tabular} & \begin{tabular}[c]{@{}c@{}}Breed \\ Distribution\end{tabular} \\ \hline
\begin{tabular}[c]{@{}c@{}}Neurologically \\ affected\end{tabular} & 2 & 2 & 1 & 1 & \begin{tabular}[c]{@{}c@{}}1 Border Terrier, 1 mixed-breed, \\ 1 French Bulldog, 1 Hovawart, \\ 1 Caucasian Ovcharka, 1 Tibetan Terrier\end{tabular} \\ \hline
\begin{tabular}[c]{@{}c@{}}Orthopedically\\ affected\end{tabular} & - & 3 & 3 & - & 4 mixed-breed dogs, 2 Fox Terrier \\ \hline
\begin{tabular}[c]{@{}c@{}}Clinically\\ healthy\\ controls\end{tabular} & 2 & 6 & 9 & - & \begin{tabular}[c]{@{}c@{}}10 Beagles, 1 mixed-breed dog, \\ 3 Fox Terriers, 1 Swiss Mountain Dog, \\ 1 Parson Russell Terrier, 1 Shar Pei\end{tabular} \\ \hline
Total & 4 & 11 & 13 & 1 & - \\ \hline
\end{tabular}
}
\captionsetup{justification=centering}
\caption{Sex distribution and breed composition.}
\label{table:breeds}
\end{table}

Table \ref{table:breeds} is presenting the sex distribution and breed composition within each group among our subjects. All subjects underwent a rigorous clinical assessment encompassing general, orthopedic, and neurological examinations by a specialist or specialist in training of the respective EBVS colleges. Furthermore, a subset of neurologically affected subjects underwent magnetic resonance imaging (MRI) to corroborate clinical findings and refine diagnostic accuracy.

Based on the clinical assessment, the three classes for our models are: {\sf healthy} dogs (no conditions diagnosed), {\sf orthopedic} dogs (diagnosed with an orthopedic issue in one of their limbs), and {\sf neurological} dogs (diagnosed with a neurological condition).

\begin{figure}[h!]
    \centering
    \captionsetup{justification=centering}
    \includegraphics[scale=0.65]{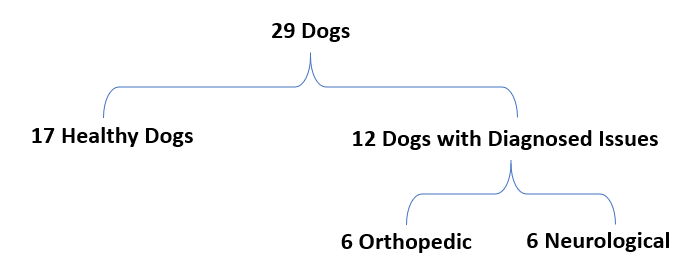}
    \caption{Number of dogs in each class.}
    \label{fig:dog_num}
\end{figure}

Figure \ref{fig:dog_num} present the number of recorded dogs in each class with a total of 29 dogs in the dataset.

With some exceptions, each dog performed two different activities: pace (walking) and trot (running). In the neurological class, only two dogs were able to trot, leading to a significant class imbalance in the trot scenarios.

The recordings of all dogs were taken on the same trajectory. It consists of three back-and-forth walks (6 passes) on a 6-meter-long walkway, (as shown in Figure \ref{fig:walking}). This results in a total trajectory length of 36 meters. However, the duration varies between dogs, due to breed differences and diseases, with some completing it faster than others. The average duration for each class, as well as the total length of the relevant data, is presented in Table \ref{table:class_duration}.

\begin{table}[h!]
\centering
\begin{tabular}{c|cc|cc|}
    \cline{2-5}
    \textbf{} & \multicolumn{2}{c|}{\textbf{Walk}} & \multicolumn{2}{c|}{\textbf{Trot}} \\ \cline{2-5}
    \textbf{} & \textbf{Avg [s]} & \textbf{Total [s]} & \textbf{Avg [s]} & \textbf{Total [s]} \\ \hline
    \multicolumn{1}{|c|}{Healthy}      & 27   & 1356  & 18  & 859   \\ \hline
    \multicolumn{1}{|c|}{Orthopedic}   & 33   & 598   & 18  & 274   \\ \hline
    \multicolumn{1}{|c|}{Neurological} & 40   & 679   & 15  & 91    \\ \hline
    \multicolumn{1}{|c|}{Total [s]}    & -    & 2633  & -   & 1224  \\ \hline
\end{tabular}
\caption{Average and total trajectory duration (in seconds) for each class and gait type.}
\label{table:class_duration}
\end{table}

As previously described, each recording includes data from three sensors placed on the dog's back and collar. Since the table presents values for a single sensor, the total recorded data is three times the values shown. In total, we collected $193$ minutes of data used for in the model, when $154$ minutes are used for training and $39$ minutes are for testing.

\subsection{Sensor Configurations}
We used the Xsens Movella DOT IMU, a compact, wearable sensor designed for high-resolution motion tracking. The DOT operates at a sampling rate of 120 Hz, which enables accurate capture of dynamic movements \cite{xsensdot}. Its accompanying software that supports time-synchronized recording across multiple IMUs, ensuring temporal consistency between sensor streams. Table~\ref{table:imu_specs} summarizes the key performance specifications of the gyroscope and accelerometer embedded in the device, including their respective bias and noise levels.

\begin{table}[h!]
\centering
    \begin{tabular}{|c|c||c|c|}
        \hline
        \multicolumn{2}{|c||}{\textbf{Gyroscope}} & \multicolumn{2}{c|}{\textbf{Accelerometer}} \\ \hline
        Bias [$^\circ$/h] & Noise [$^\circ$/s/$\sqrt{\text{Hz}}$] & Bias [mg] & Noise [$\mu$g/$\sqrt{\text{Hz}}$] \\ \hline
        10 & 0.007 & 0.03 & 120 \\ \hline
    \end{tabular}
\captionsetup{justification=centering}
\caption{Movella DOT IMU sensor specifications.}
\label{table:imu_specs}
\end{table}

\begin{figure}[H]
    \centering
    \captionsetup{justification=centering}
    \includegraphics[scale=0.5]{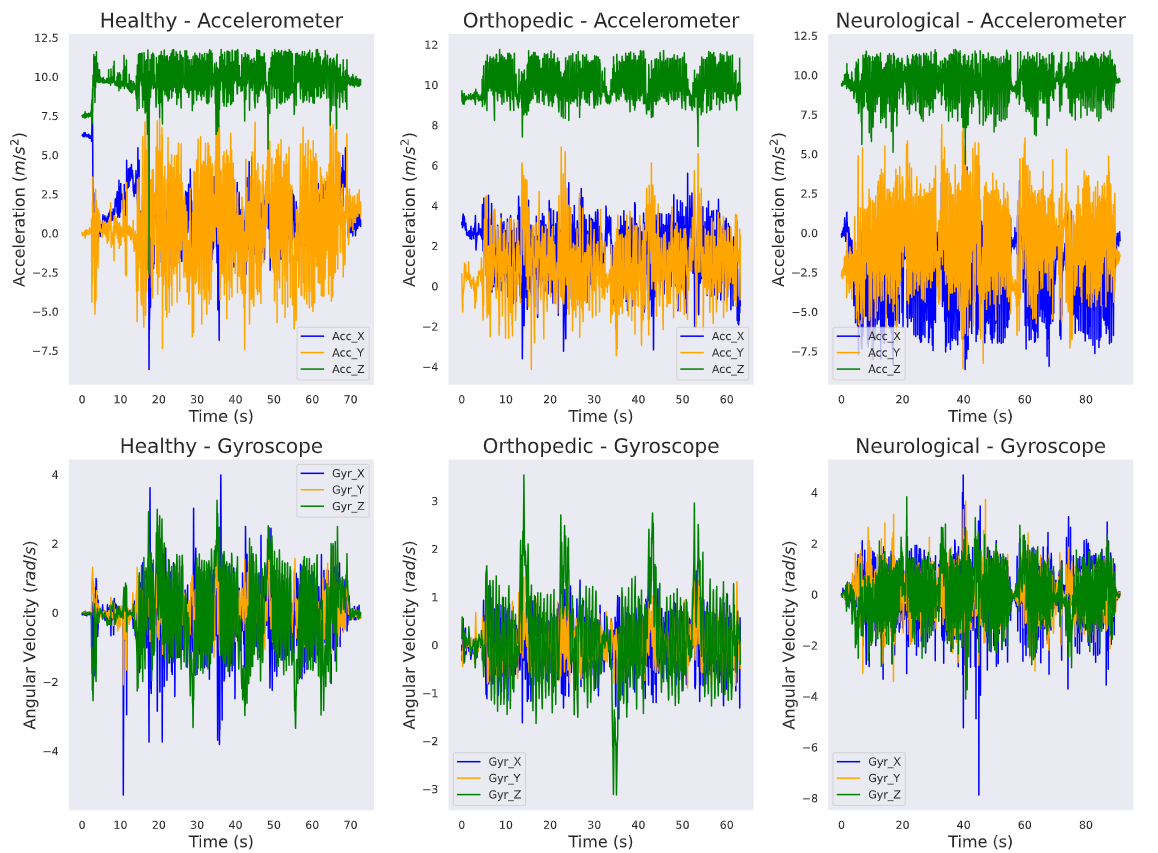}
    \caption{Raw IMU data of the different axes for the 3 classes.}
    \label{fig:raw_data}
\end{figure}

Figure \ref{fig:raw_data} presents an example of the inertial raw data for healthy, orthopedic, and neurological dogs.

\begin{figure}[h!]
    \centering
    \captionsetup{justification=centering}
    \includegraphics[scale=0.42]{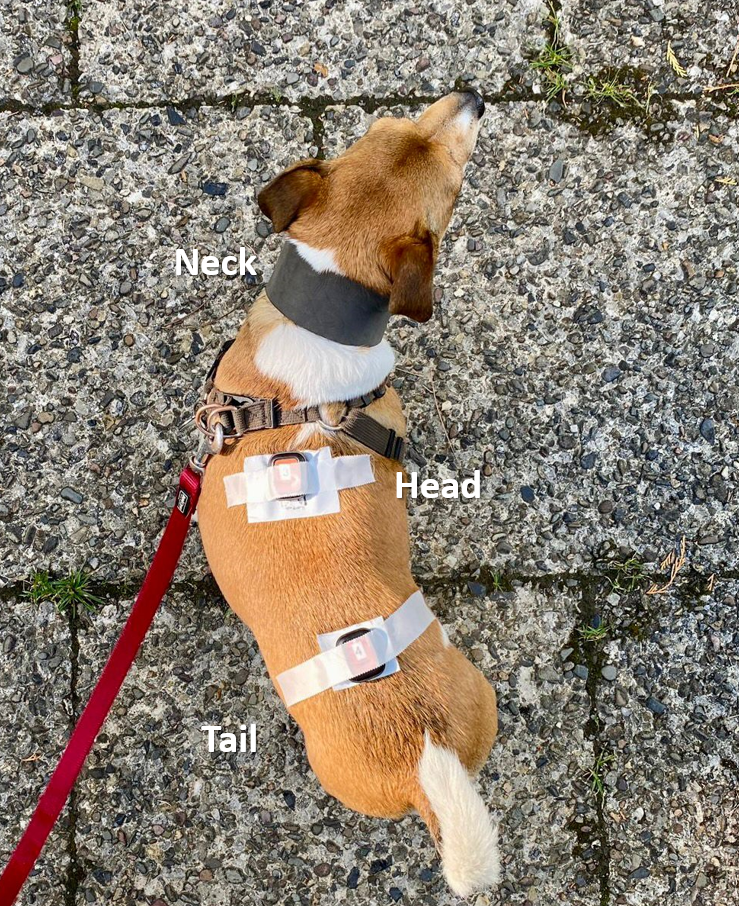}
    \caption{Placement of the IMUs on the dog.}
    \label{fig:sensors_dog}
\end{figure}

Figure \ref{fig:sensors_dog} illustrates the locations for placing the sensors chosen for minimization of discomfort for the tested dogs. We denote the three different locations of the sensors on the dog as follows: 1) Head - Top back, 2) Tail - Lower back, 3) Neck - Collar. Thus, for each dog, six different combinations of sensor location and walking type were tested.

The raw data was preprocessed, filtering noise (e.g., prior to the dog's walking) by analyzing the norm of the accelerometer measurements and determining a fixed window and stride size, which allows the data signal to be analyzed as a multivariate time series. We experimented with different window sizes and strides to find the optimal configuration. After a prior analysis we chose a window size of 120 and a stride of 5.

\section{Analysis and Results} \label{results}

We began our analysis by evaluating the performance of our approach using a known group of dogs that participated in our data collection process (baseline). Next, we examined the generalizability of our approach by extending the evaluations to unseen dogs. Lastly, we applied various techniques to improve the performance of our generalized model. In this section, we detail these steps, present the corresponding results, and summarize the most effective approaches based on the obtained outcomes.

\subsection{Baseline Models} \label{radnom results}

The first scenario focused on a closely monitored group of dogs without the need to generalize to additional dogs outside this group. As such, it involved a train/test approach based on a random split of the dataset.
In this setup, we utilized the previously described network architecture (Figure \ref{fig:archi}) and applied it across the three classification scenarios: Multi, Binary, and Diagnosis, as described in Section \ref{methodology}. The network architecture remained identical in all cases, with the only difference being the number of output units in the final layer - set to either two or three, corresponding to the classification task. The metrics we used to evaluate performance were accuracy and F1 score.

\subsubsection{Binary Classification}

\begin{figure}[h!]
\centering
    \captionsetup{justification=centering}
    \includegraphics[scale=0.6]{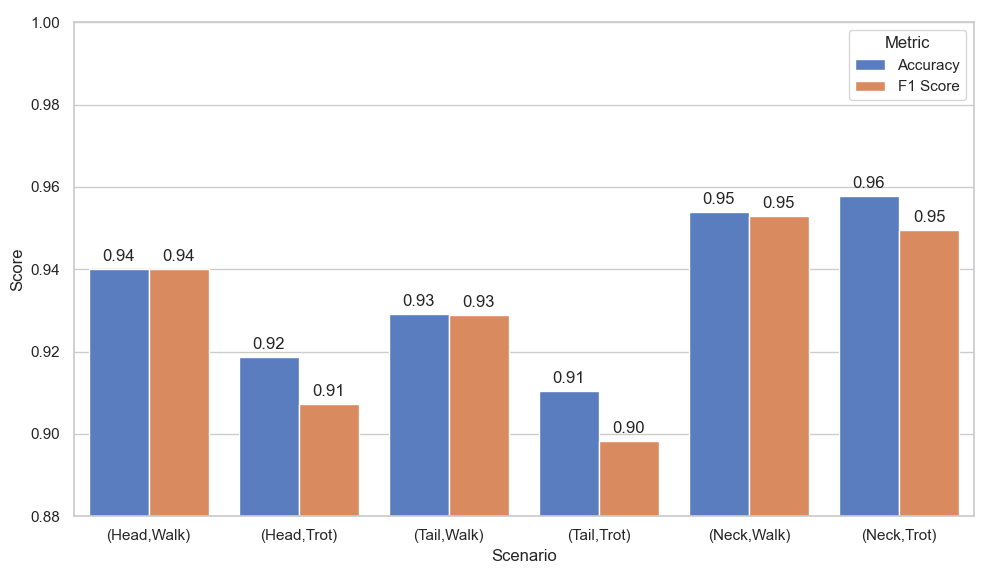}
    \caption{Binary classification accuracy and F1 results across sensor locations and protocols.}
\label{fig:HS_random}
\end{figure}

This scenario aimed to answer the following question: could we distinguish between healthy individuals and those exhibiting signs of illness? Figure \ref{fig:HS_random} presents the classification performance for the binary classification task across the different walking protocols and sensor locations. The results showed consistently high accuracy and F1 scores, with the best performance observed using the Neck sensor location, regardless of protocol type (walk/trot), achieving 0.95–0.96 accuracy and F1 scores.

\vspace{6mm}
\subsubsection{Diagnosis Classification}

\begin{figure}[h!]
\centering
    \captionsetup{justification=centering}
    \includegraphics[scale=0.55]{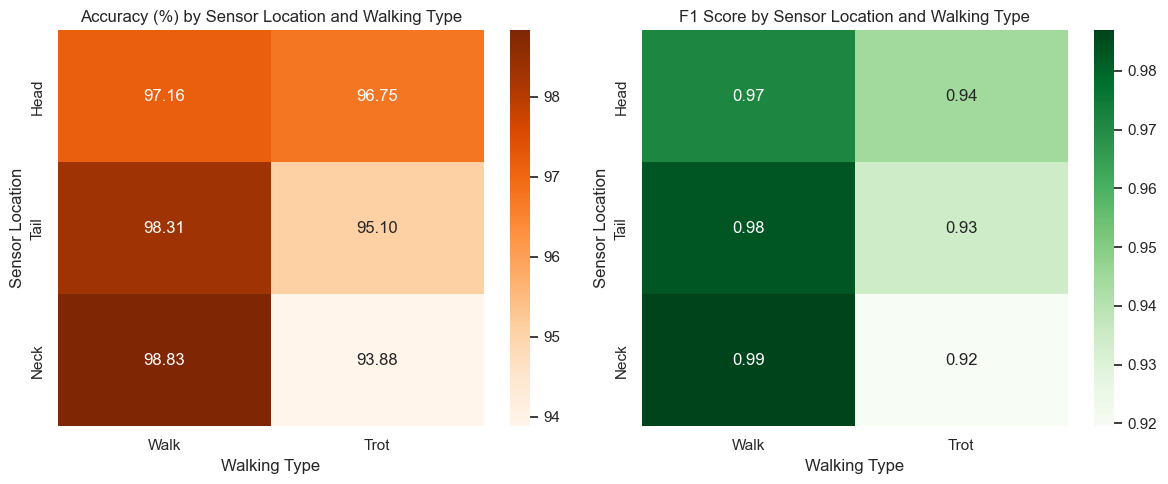}
    \caption{Diagnostic classification performance.}
\label{fig:ON_random}
\end{figure}

In this scenario, we focused exclusively on dogs diagnosed with an illness and aimed to differentiate between orthopedic and neurological issues. Figure \ref{fig:ON_random} presents the model’s performance on the orthopedic vs. neurological classification task across various sensor locations and protocols. The best performance was observed at the Neck location while walking, achieving 98.83\% accuracy and an F1 score of 0.99. However, a consistent drop in both metrics was seen when switching to trotting, particularly at the Neck location, where accuracy and F1 score decreased to 93.88\% and 0.92, respectively. These results suggested that the protocol significantly influenced classification performance and highlighted walking as the more reliable protocol.

\subsubsection{Multiclass Classification}

In this scenario, we aimed to answer the following question: given a group of dogs without prior diagnostic information, could we accurately determine whether each dog was healthy, or if not, whether the underlying issue was orthopedic or neurological? Table \ref{table:rand_res_mul} presents the performance of the baseline model using sensor data as input in the multiclass classification task. The results showed consistently high accuracy and F1 scores, reaching up to 98.38\% accuracy and a 0.97 F1 score (trotting protocol with the Neck sensor). Even the lowest-performing configuration (walking protocol with the Tail sensor) still achieved 93.18\% accuracy, demonstrating the model's robustness across settings.

\begin{table}[h!]
\centering
    \begin{tabular}{|c|c|c|c|}
    \hline
    \textbf{Sensor Placement} & \textbf{Protocol} & \textbf{Accuracy} & \textbf{F1} \\ \hline
    \multirow{2}{*}{Head}     & Walk              & 94.97             & 0.94        \\ \cline{2-4} 
                              & Trot              & 96.3              & 0.96        \\ \hline
    \multirow{2}{*}{Tail}     & Walk              & 93.18             & 0.92        \\ \cline{2-4} 
                              & Trot              & 96.84             & 0.97        \\ \hline
    \multirow{2}{*}{Neck}     & Walk              & 96.55             & 0.96        \\ \cline{2-4} 
                              & Trot              & 98.38             & 0.97        \\ \hline
    \end{tabular}
\captionsetup{justification=centering}
\caption{Multiclass classification performance results across sensor configurations: random split.}
\label{table:rand_res_mul}
\end{table}

\begin{figure}[h!]
  \centering
  \begin{subfigure}[b]{0.45\textwidth}
    \centering
    \includegraphics[width=0.71\textwidth]{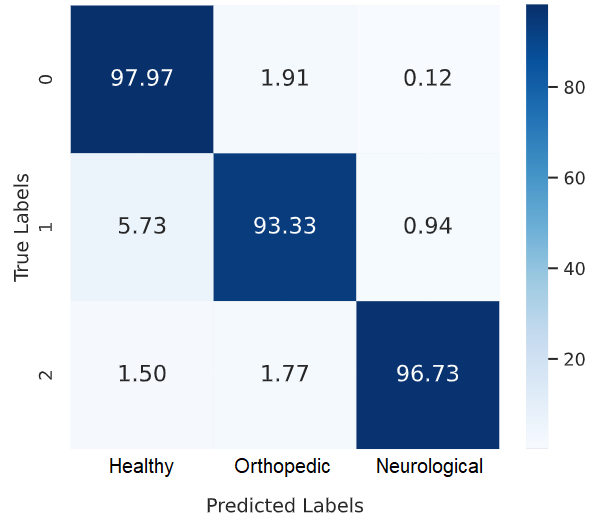}
    \caption{\parbox[t]{.85\linewidth}{Confusion matrix for (Neck, walking).}}
    \label{fig:conMatP}
  \end{subfigure}
  \hspace{0.02\textwidth}
  \begin{subfigure}[b]{0.45\textwidth}
    \centering
    \includegraphics[width=0.71\textwidth]{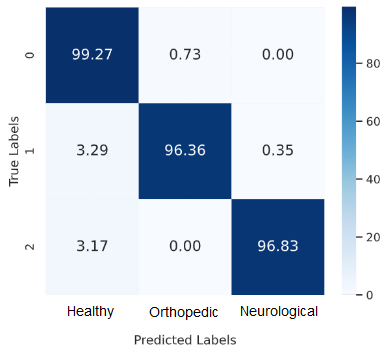}
    \caption{\parbox[t]{.85\linewidth}{Confusion matrix for (Neck, trotting).}}
    \label{fig:conMatT}
  \end{subfigure}
  \caption{Confusion matrices for Neck sensor placement: (a) walking, (b) trotting.}
  \label{fig:conMat}
\end{figure}

Figure \ref{fig:conMat} provides further insight into multiclass classification performance by showing the confusion matrices for the best-performing sensor placement (Neck). In both walking and trotting, the model demonstrated strong classification ability with minimal confusion between classes. Misclassifications were rare and occurred primarily between the orthopedic and neurological classes, which were naturally more similar.

\subsection{Generalization} {\label{gen}}

Here, we addressed the generalization issue - specifically, whether the results could be extended to any unseen dog not included in the training dataset. To this end, we applied a Leave-One-Out (LOO) train/test split approach. In each iteration, the model was trained on all dogs except one, which was then used for testing. This process was repeated for each dog, and accuracy and F1 scores were recorded for every run. The reported results represent the mean performance across all iterations. As an initial step in this approach, we performed the same evaluations as those conducted for the closely monitored group, as presented in Section \ref{radnom results}.

\subsubsection{Binary Classification}
\begin{figure}[h!]
\centering
    \captionsetup{justification=centering}
    \includegraphics[scale=0.5]{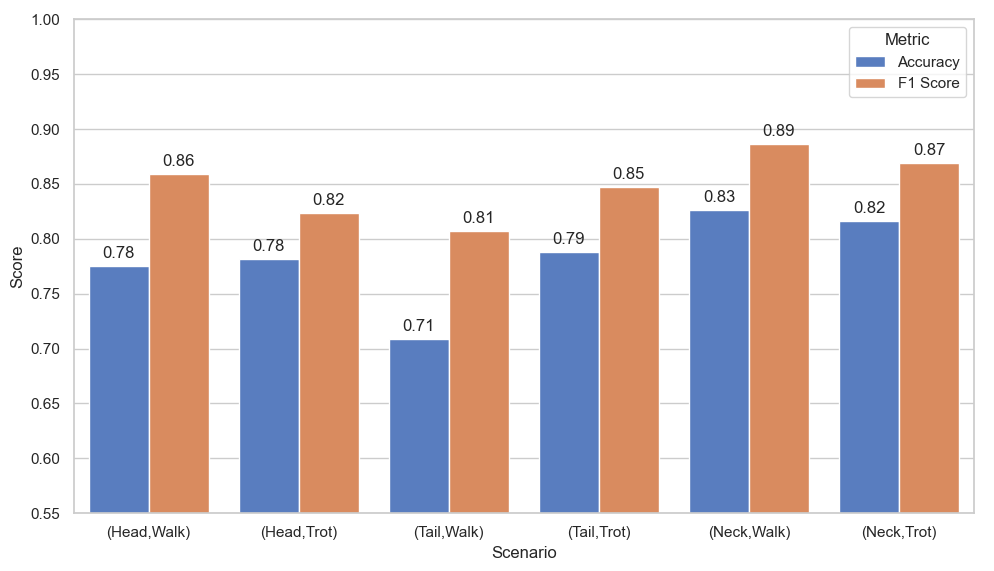}
    \caption{Binary classification F1 and accuracy results for our model generalization.}
\label{fig:stat_graph_hs}
\end{figure}

Figure \ref{fig:stat_graph_hs} presents the classification performance for distinguishing between healthy dogs and those with diagnosed issues under the different scenarios. Overall, the best performance was observed with the neck sensor, both while walking and trotting, achieving the highest F1 scores of 0.89 and 0.87, respectively. This indicated that the collar-mounted sensor was a strong choice. In contrast, the walking protocol with the tail sensor showed the lowest performance, with an accuracy of 0.71, suggesting that this combination posed greater challenges for classification. Across all scenarios, F1 scores consistently exceeded accuracy scores, emphasizing the model’s robustness in handling class imbalance and varying class distributions.

\subsubsection{Diagnosis Classification}
\begin{figure}[h!]
\centering
    \captionsetup{justification=centering}
    \includegraphics[scale=0.45]{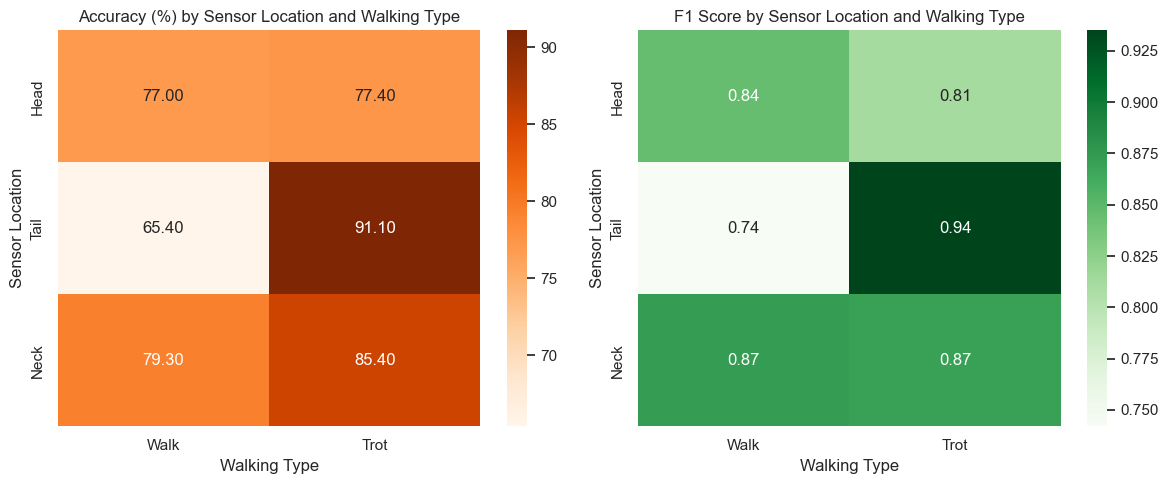}
    \caption{Diagnosis classification results for generalization.}
\label{fig:stat_graph_on}
\end{figure}

Figure \ref{fig:stat_graph_on} presents the classification performance for distinguishing orthopedic from neurological cases across the three sensor locations and two walking types. The best performance was achieved using the tail sensor during the trotting protocol, with an accuracy of 91.1\% and an F1 score of 0.94. In contrast, the lowest performance was also observed with the tail sensor but during the walking protocol, yielding only 65.4\% accuracy and an F1 score of 0.74. This indicated a substantial variation in performance for this sensor location depending on the walking type. While the trotting protocol appeared to benefit the tail sensor the most, the walking protocol provided more consistent results across the head and neck sensor placements. Overall, these findings suggested that both sensor placement and walking condition significantly influenced classification effectiveness when differentiating between orthopedic and neurological conditions.

\subsubsection{Multiclass Classification}

Table \ref{table:stat_res_hon} shows the multiclass classification results for generalization across six scenarios that combined walking types and sensor placements. The best overall performance was achieved using the neck sensor during the trotting protocol, with 78.8\% accuracy and an F1 score of 0.826. In contrast, the lowest performance was observed with the tail sensor during the walking protocol, yielding only 58.1\% accuracy and an F1 score of 0.668.

\begin{table}[h!]
\centering
    \begin{tabular}{|c|c|c|c|}
    \hline
    \textbf{Sensor Placement} & \textbf{Protocol} & \textbf{Accuracy} & \textbf{F1} \\ \hline
    \multirow{2}{*}{Head}     & Walk              & 67             & 0.768       \\ \cline{2-4} 
                              & Trot              & 74.1              & 0.767        \\ \hline
    \multirow{2}{*}{Tail}     & Walk              & 58.1             & 0.668       \\ \cline{2-4} 
                              & Trot              & 73.3             & 0.785        \\ \hline
    \multirow{2}{*}{Neck}     & Walk              & 75.7             & 0.835       \\ \cline{2-4} 
                              & Trot              & 78.8             & 0.826        \\ \hline
    \end{tabular}
\captionsetup{justification=centering}
\caption{Multiclass classification performance results across sensor configurations for the generalization scenario.}
\label{table:stat_res_hon}
\end{table}

\subsection{Performance Enhancement}

During data collection, we observed that some dogs were unable to trot (most of those are impaired dogs), so we focused on the pacing rate as our baseline and combined three different approaches to improve the network generalization. Those include:

\begin{enumerate}
    \item \textbf{Data Augmentation:} One of the biggest challenges in deep learning models is their requirement for large amounts of data, which is particularly difficult to collect in our case. One way to address this problem is by generating more data based on the data we have already collected \cite{fekson2025enhancement, yampolsky2025neural}. As part of our process, we explored different augmentation methods and scales to expand our dataset and improve generalization. The best results in this regard were achieved using a rotation matrix for the x-axis with a 15-degree angle.
    
    \item \textbf{Multi-Head Architecture:} Multi-head approaches have shown better performance compared to a single head architecture \cite{silva2019end, liu2023smartphone}. The basic idea is to process separately the accelerometer and gyroscope readings. To that end, based on our single head architecture presented in Section \ref{Net_Arch}, we define a multi-head architecture as presented in Figure \ref{fig:archi2head}.
    
    \begin{figure}[h!]
    \centering
        \captionsetup{justification=centering}
        \includegraphics[scale=0.3]{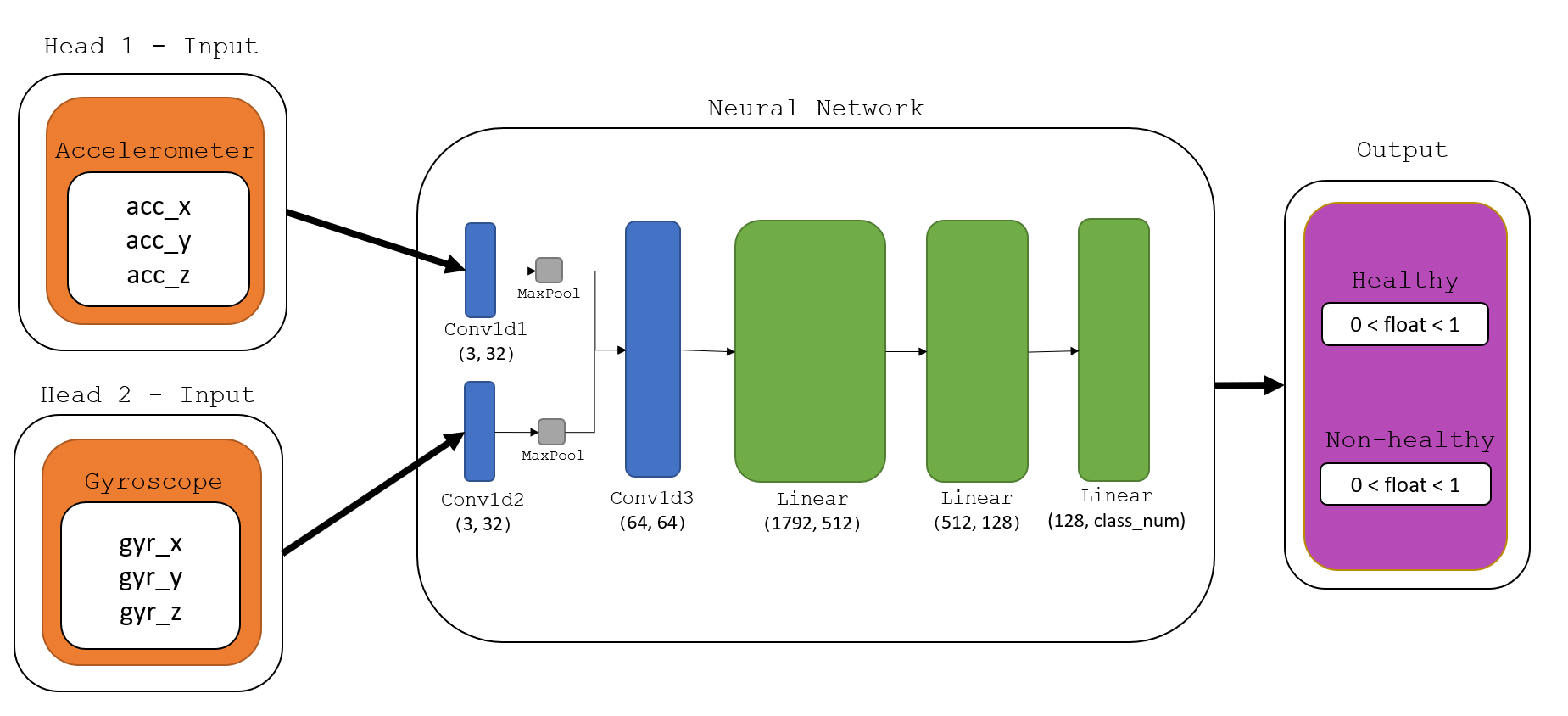}
        \caption{Two-Head architecture - one head receives accelerometer readings and the other receives gyroscope readings for the Binary classification scenario.}
        \label{fig:archi2head}
    \end{figure}
    
    \item \textbf{Sensor Location:} The available sensor location options remain the same as before, Head, Tail, and Neck, with the addition of \textbf{All} – Data from all sensor locations are merged into a single dataset used to train the model.
    
\end{enumerate}

In Table \ref{table:stat_res2}, we present the results for the Binary classification when the dogs are walking, across all the different sensor locations. These results were obtained using both data augmentation and the 2-Head approach.

\begin{table}[h!]
\centering
\captionsetup{justification=centerlast}
\begin{tabular}{|c|c|c|c|c|}
\hline
\makecell{\textbf{Sensor} \\ \textbf{Placement}} &
\makecell{\textbf{Accuracy} \\ \textbf{(\%)}} &
\makecell{\textbf{$\Delta$ Accuracy} \\ \textbf{(\%)}} &
\makecell{\textbf{F1} \\ \textbf{Score}} &
\makecell{\textbf{$\Delta$ F1} \\ \textbf{Score}} \\ \hline
Head                      & 82                     & \textcolor{green}{$\uparrow$ 5}     & 0.900             & \textcolor{green}{$\uparrow$ 0.04} \\ \hline
Tail                      & 77.1                   & \textcolor{green}{$\uparrow$ 6}     & 0.859             & \textcolor{green}{$\uparrow$ 0.05} \\ \hline
Neck                      & 80.3                   & \textcolor{red}{$\downarrow$ 2}       & 0.863             & \textcolor{red}{$\downarrow$ 0.025}  \\ \hline
All                       & 77.9                   & –                      & 0.871             & –                      \\ \hline
\end{tabular}
\caption{Performance after data augmentation with 2-Head approach\\ (Binary classification, walking)}
\label{table:stat_res2}
\end{table}

The results are quite similar across the different configurations, but it appears that the best model performance is achieved when using the top back sensor (Head).
Interestingly, the augmentation and two-head approach appear to reduce the performance for the Neck sensor location, in contrast to the other two sensor locations, which show an improvement of ~5\% in accuracy. This suggests that these methods are not universally beneficial across all scenarios and may require further investigation to understand their limitations.

\subsection{Summary}
Table \ref{table:FULL_RES} summarizes the most significant results of this research. It is important to emphasize again that some dogs were unable to trot or could not wear the collar-mounted sensor. Therefore, while identifying the configurations that yield the best results is valuable, it is equally crucial to balance performance with practicality-ensuring that every dog can benefit from a simple, accessible, and efficient model.
In the closely monitored group approach, results across all cases are consistently strong and comparable. Therefore, the most reliable and convenient choice is the multi-class classification using a single back-mounted sensor (Head), which achieves excellent performance with 95\% accuracy and an F1 score of 0.94.

\begin{table}[h!]
\centering
\resizebox{\textwidth}{!}{%
\begin{tabular}{|c|c|c|c|c|c|}
\hline
\textbf{Classification Task}                                                    & \textbf{Setting} & \textbf{\begin{tabular}[c]{@{}c@{}}Best\\ Sensor Placement\end{tabular}} & \textbf{\begin{tabular}[c]{@{}c@{}}Best\\ Walking Protocol\end{tabular}} & \textbf{\begin{tabular}[c]{@{}c@{}}Accuracy\\ (\%)\end{tabular}} & \textbf{\begin{tabular}[c]{@{}c@{}}F1\\ Score\end{tabular}} \\ \hline
Binary                                                                & Baseline        & Neck                                                                    & Trot                                                                  & 96                                                               & 0.95                                                        \\ \hline
Binary                                                                & Generalization   & Head                                                                    & Walk                                                                  & 82                                                               & 0.9                                                         \\ \hline
Diagnosis                                                     & Baseline        & Neck                                                                    & Walk                                                                  & 98                                                               & 0.99                                                        \\ \hline
Diagnosis                                                     & Generalization   & Tail                                                                    & Trot                                                                  & 91                                                               & 0.94                                                        \\ \hline
Multi-class     
        & Baseline        & Neck                                                                    & Trot                                                                  & 98                                                               & 0.97                                                        \\ \hline
Multi-class
        & Generalization   & Neck                                                                    & Trot                                                                  & 79                                                               & 0.82                                                        \\ \hline
\end{tabular}
}
\caption{Summary of the significant results of this research.}
\label{table:FULL_RES}
\end{table}

\section{Conclusions} \label{conclusions}
The results of this study demonstrate the strong potential of our model to classify non-healthy dogs from healthy ones, as well as to distinguish whether the cause of the problem is neurological or orthopedic. We can extract sufficient insights into the dog's gait patterns by using a single sensor placed comfortably on the upper back or neck of the dog - based on whether the dog tolerates wearing a collar-mounted device - to collect inertial readings (accelerometer and gyroscope) while walking the dog on a straight path.

In the Baseline case, where the model is applied to a known and fixed group of dogs, it achieves high performance across all tasks, with up to 96\% accuracy in the Multi-class classification. This setting is simplifies clinical decision-making by reducing the need for extensive diagnostic procedures.

In the more challenging Generalization case, where the model is evaluated on new, unseen dogs, it still shows strong results: achieving 82\% accuracy in the Binary classification (healthy vs. non-healthy) and 77\% accuracy in the Diagnosis classification (orthopedic vs. neurological). This demonstrates the model’s potential as a practical diagnostic support tool.

Overall, this non-invasive approach - using a single wearable sensor and requiring only natural walking - can help veterinarians detect and distinguish between gait-related pathologies early, offering more targeted and timely treatment.

Several additional evaluations can be conducted to enhance the model's performance further. For example, our model presented interesting results using the Neck sensor; however, two dogs were unable to wear a collar-mounted sensor. Another important area for improvement is the trotting data, which has shown to enhance classification accuracy, but is usually not feasible with neurological impaired dogs. Nevertheless, it offers a great opportunity to evaluate and monitor orthopedically impaired dogs whenever possible. Thus, while our model shows promising results, it also highlights the potential for further exploration into different sensor placements, gait types, and data augmentation techniques, which could lead to even better performance and broader clinical applicability.

In conclusion, this work provides a novel, non-invasive approach for the early detection and classification of gait abnormalities in dogs, with the potential to significantly enhance diagnostic accuracy in veterinary practice. By leveraging wearable inertial sensors and deep learning models, the proposed method offers an efficient, cost-effective, and accessible tool for identifying orthopedic and neurological conditions. This work stands to benefit clinical veterinarians, researchers in animal biomechanics, and dog owners alike - contributing to earlier intervention, more informed treatment decisions, and ultimately, improved welfare and quality of life for canine patients.

\section*{Acknowledgement}
N. P. was supported by the Maurice Hatter Foundation.

\bibliographystyle{ieeetr}
\bibliography{Ref}

\end{document}